%% file: root.tex
\documentclass[a4paper,conference]{IEEEtran}

\usepackage{graphicx}
\usepackage{comment}
\usepackage{enumitem}
\usepackage[dvipsnames]{xcolor}
\usepackage{url}
\usepackage{booktabs}
\usepackage{multicol}

\input{macros}

\begin{document}

\title{\emph{Transforming} Image Generation from Scene Graphs}

\author{
 \IEEEauthorblockN{Renato Sortino, Simone Palazzo, Concetto Spampinato}
 \IEEEauthorblockA{PeRCeiVe Lab, \\
                   University of Catania, Italy.\\
                   \url{www.perceivelab.com}
 }
}

\maketitle

\begin{abstract}
\input{Sections/0_Abstract}
\end{abstract}

\section{Introduction}
\label{sec:introduction}
\input{Sections/1_Intro}

\section{Related work}
\label{sec:related}
\input{Sections/2_Related_Work}

\section{Method}
\label{sec:method}
\input{Sections/3_Method}

\section{Performance Analysis}
\label{sec:results}
\input{Sections/4_Results}

\section{Conclusion}
\label{sec:conclusion}
\input{Sections/5_Conclusions}

\section*{Acknowledgements}
\label{sec:acknowledgements}
\input{Sections/6_Acknowledgements}

\bibliographystyle{IEEEtran}
\bibliography{root.bib}

\end{document}

%% file: macros.tex
\usepackage{pifont}
\usepackage{amsmath}
\usepackage{amssymb}
\usepackage{color,soul}

\newcommand{\mytilde}{\raise.17ex\hbox{$\scriptstyle\mathtt{\sim}$}}

%% file: Sections/0_Abstract.tex
Generating images from semantic visual knowledge is a challenging task, that can be useful to condition the synthesis process in complex, subtle, and unambiguous ways, compared to alternatives such as class labels or text descriptions.
Although generative methods conditioned by semantic representations exist, they do not provide a way to control the generation process aside from the specification of constraints between objects. As an example, the possibility to iteratively generate or modify images by manually adding specific items is a desired property that, to our knowledge, has not been fully investigated in the literature. 
In this work we propose a transformer-based approach conditioned by scene graphs that, conversely to recent transformer-based methods, also employs a decoder to autoregressively compose images, making the synthesis process more effective and controllable.
The proposed architecture is composed by three modules: 1) a graph convolutional network, to encode the relationships of the input graph; 2) an encoder-decoder transformer, which autoregressively composes the output image; 3) an auto-encoder, employed to generate representations used as input/output of each generation step by the transformer. Results obtained on CIFAR10 and MNIST images show that our model is able to satisfy semantic constraints defined by a scene graph and to model relations between visual objects in the scene by taking into account a user-provided partial rendering of the desired target.

%% file: Sections/1_Intro.tex
The investigation of learning-based approaches for image generation has been an active research topic in the last decade, in particular since the introduction of Generative Adversarial Networks (GANs)~\cite{goodfellow2014generative}, which have paved the way to a whole family of techniques for photorealistic image synthesis~\cite{dcgan,biggan,stylegan}.

One of the most attractive features of GANs is their capability to learn realistic visual patterns in an unsupervised way from a large enough set of unlabeled images, making them suitable to support model training in semi-supervised scenarios~\cite{gan-semisupervised}, or as a data augmentation technique~\cite{pennisi2021self,gan-data-augmentation-1}. However, being able to control the synthesis process, at the cost of providing additional metadata describing the properties of the desired output, is a key factor for certain applications, including image translation~\cite{pix2pix}, user-controlled CGI~\cite{vid2vid,stylegan}, or in domain-specific scenarios with low data availability~\cite{gan-data-augmentation-2}. The type of control that can be integrated in such models may vary, depending on the strength of the prior that should be enforced: class label conditioning~\cite{cgan} is the most straightforward option, when specific visual characteristics are not required. Conditioning through a set of discrete learned embeddings has also been proposed~\cite{vid2vid}, allowing users to choose between a set of styles. A less constraining and more intuitive modality consists in employing text descriptions as a way to condition a model~\cite{text-to-image}; this kind of approaches, while being more user-friendly, may produce unexpected results due to possible ambiguity in the description, or to limited understanding capabilities of the language model.

Structured semantic representations, such as scene graphs~\cite{johnson2018image} or computational ontologies~\cite{paper-ontologies}, represent the ideal choice for visual constraint description applied to image generation, as they support the definition of well-defined, arbitrary and unambiguous sets of entities and corresponding relations. These representations are also suitable for tasks at different levels of complexity, from simple ones (e.g., where the relative positions of common-life objects are described) to hard ones (e.g., in domain-specific scenarios, as in~\cite{paper-ontologies}). Current approaches for image generation using graph-based representations generally combine graph convolutions for constraint analysis and convolutional neural networks for output synthesis~\cite{johnson2018image}. One of the downsides of such pipeline is that the conditioning is limited to providing scene constraints as an input to the model, which is otherwise free to synthesize objects arbitrarily, as long as they satisfy the structure defined by the graph. As a consequence, it is not trivial to either 1) model the generation process as a sequence of steps (e.g., each one associated to an object in the scene), or 2) condition the generator with a partial rendering of the desired output.

In this work, we present an approach that aims to overcome these limitations by posing the image generation task as an autoregressive one, where a model gradually adds scene elements that satisfy the constraints defined by a scene graph. The ``starting point'' of the generation process can be either an empty image, in which case the model has to fill it with \emph{all} entities described by the scene graph specification, or a partial rendering of the target scene: in this more general case, the model \emph{completes} the input scene by synthesizing only the elements needed to satisfy the constraints. As a result, this approach implements a strategy to further control the generation process by giving users the possibility to provide an initial image layout partially filled with objects, and let the model complete it.

Our approach is based on a combination of a graph convolutional network (GCN)~\cite{gcn}, an autoencoder as image descriptor~\cite{autoencoder}, and an encoder-decoder transformer~\cite{transformer} for synthesis. In particular, the GCN first processes the scene graph provided as input, by extracting features that are fed to the transformer encoder. 
Then, the transformer decoder autoregressively builds the desired scene, by receiving an initial \emph{start-of-sequence} input --- which can be either a blank scene or an initial rendering of the target scene. At each generation step, the transformer decoder inserts new elements in the scene: it receives image features (extracted by the autoencoder) corresponding to the previously-generated partial scenes, and produces an image representation that is converted (again through the autoencoder) into the new generated image.

Performance analysis, carried out on a customization of MNIST and CIFAR10 datasets that include scene graphs describing simple layout constraints, shows that our method promisingly generates images consistently meeting  the scene graph specifications, in line with state of the art methods such as \emph{sg2im}~\cite{johnson2018image}. However, to the best of our knowledge, our approach is the first that is able to condition the generation process by taking into account a combination of semantic knowledge and user-provided partial rendering of the desired target. 

%% file: Sections/2_Related_Work.tex
In recent years, the image generation task has been tackled using mainly Generative Adversarial Networks~\cite{goodfellow2014generative,biggan,stylegan}, whose success has led to their employment in multiple tasks, such as data augmentation~\cite{sundaram2021ganbased}, super-resolution~\cite{zhang2021ranksrgan},
image denoising~\cite{huang2021dugan} and domain adaptation~\cite{huang2021generation}. The idea underlying the GAN approach is to match a dataset distribution by training a discriminator model to distinguish between samples from and outside of the distribution, and a generator model to produce outputs that are interpreted by the discriminator as belonging to the distribution.

The basic GAN formulation, however, does not provide a direct way to control the generated output. A first step in this direction is introduced by conditional GANs~\cite{cgan}, where the generator and discriminator are provided with additional information encoded as a class label embedding or representing, in general, a latent space that describes specific visual properties. 

A particularly interesting way for conditioning GAN models is through text descriptions of the expected image content, thus providing a high degree of flexibility to the representation, as well as an intuitive way for users to interact with the model. One of the first approaches devised to tackle text conditioning was presented in \cite{text-to-image}, where learned word embeddings are used to control the generation process. Later approaches follow this idea by integrating architectural improvements to GAN models, such as stacked GANs and Hierarchical GANs~\cite{zhang2017stackgan,zhang2018stackgan,zhang2018photographic}.
Further refinements in these approaches apply attention mechanisms to improve the representation used for text embedding, by capturing the relationships between words within a sentence and visual patches: some examples of this approach are AttnGAN~\cite{xu2017attngan} and ControllableGAN~\cite{li2019control}.

Since the diffusion of the transformer architecture \cite{transformer} and its increasing employment for vision tasks~\cite{vit,detr,transformer-segmentation,timesformer}, new generative approaches based on self-attention feature learning have been proposed. GANformer~\cite{ganformer} employs a bipartite structure that enables long-range interactions and iteratively propagates information from a set of latent variables to the evolving visual features and vice versa, to encourage the emergence of compositional representations of objects and scenes. A text-to-image generation approach is presented in~\cite{ramesh2021zeroshot}, where a discrete variational autoencoder is used to compress images and a decoder-only transformer regresses visual codebooks from text embeddings. This work shows that transformers, along with a high amount of data and training resources, are capable of encoding abstract concepts and synthesize them together in the visual space.

While text to image is widely covered and experimented, image generation conditioned on scene graphs has not been yet exhaustively explored.
The first work which analyzed an architecture based on structured semantic conditioned image generation is~\cite{johnson2018image}, that processes the input scene graphs with a GCN,  projects the learned embedding into object latent vectors and image layouts, and finally renders the output image through a GAN. PasteGAN~\cite{li2019pastegan} improves the image generation process by using an additional conditioning of the GAN. For each object, this model selects a crop from an external memory, i.e., the one whose features better match the encoded input scene graph, and uses it to condition the generation.  Following this line of research, more recent approaches attempted to improve the generation process, by processing \emph{subject-predicate-object} relations individually~\cite{vo2020visualrelation}, reducing predicates to a canonical form~\cite{canonsg2im}, or iteratively modifying the scene graph while retaining previously-generated content~\cite{mittal2019interactive}.

Similarly to the these methods, our work processes scene graphs through a GCN in order to extract a representation that encodes object appearance and layout. However, we introduce a novel method to condition the generation of the output from partial renderings of the desired target, by employing an encoder-decoder transformer architecture that supports intermediate outputs as start-of-sequence for the decoder, and completes the generation by filling in the remaining elements defined in the graph. Compared to~\cite{johnson2018image},~\cite{li2019pastegan},~\cite{mittal2019interactive},~\cite{canonsg2im} our approach provides an additional conditional mechanism besides the scene graph, i.e., the intermediate desired output; this is a more direct conditioning than scene graph modifications, enabling the possibility to control the generation process from the very beginning of synthesis.

%% file: Sections/3_Method.tex
\begin{figure*}
	\centering
	\includegraphics[width=0.80\textwidth]{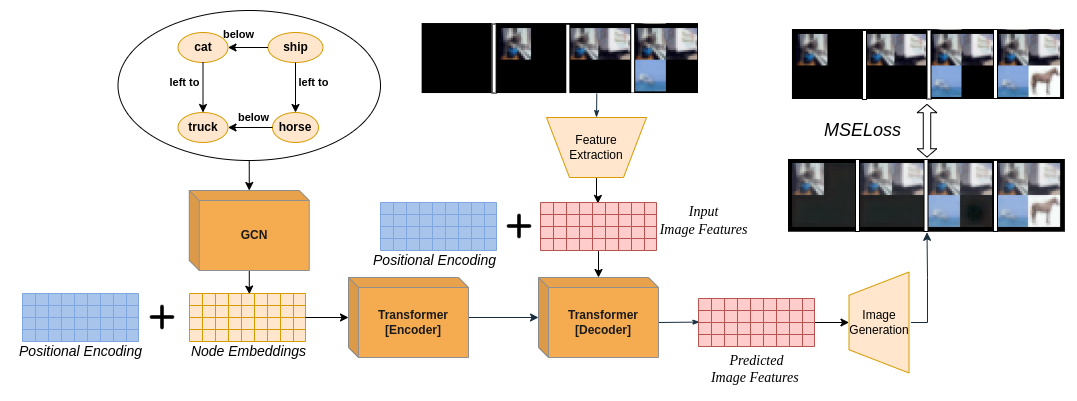}
	\caption{\textbf{Overall model architecture.} From left to right, the scene graph is provided as input to the GCN and the produced node embeddings are combined with a positional encoding and then fed to the transformer encoder. At each decoding step, the transformer receives features extracted from the current partial image and produces new image features, which are projected onto the visual space by the image generator. An MSE loss function on each decoder step output is employed to train the model to gradually build the target output image.}
	\label{fig:architecture}
\end{figure*}
\begin{figure}
	\centering
	\includegraphics[width=0.35\textwidth]{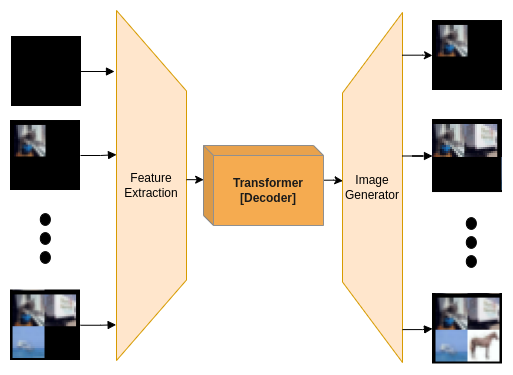}
	\caption{\textbf{Autoregressive image generation.} At each step, the decoder receives the previously generated image and produces a new image which includes a new element in the scene. The newly generated image is then provided at the input of the next iteration, until all elements from the target scene graph have been produced. The starting point of the generation can be either a blank image or a partial rendering of the target.}
	\label{fig:decoder}
\end{figure}

The proposed approach can be decomposed in terms of its three key features: 1) modeling scene graphs through graph convolutional networks
for conditioning the generation process; 2)
using the graph embeddings as input to an encoder-decoder transformer architecture that auto-regressively models semantic constraints and visual elements; and 3) learning an efficient continuous representation for transformer visual ``tokens'' by means of an autoencoder trained on the image dataset, to be used as feature extractor and image generator. The integrated architecture of our approach is shown in Fig.~\ref{fig:architecture}, with the detail of the auto-regressive generative process illustrated in Fig.~\ref{fig:decoder}.

\subsection{Scene Graph Modeling}

We employ a graph convolutional network (GCN) to process the input scene graph, by extracting features for each node based on the neighbouring nodes, their relationships (defined as edge features), and the overall graph structure.

The core mechanism of the GCN can be formally described as a message passing network~\cite{gilmer2017neural}, which updates each node state by aggregating neighbouring node states with a permutation-invariant function, such as sum or mean:
\begin{equation}
    h_{v}^{(t+1)} = U_t\left(h_{v}^{(t)}, m_{v}^{(t+1)}\right),
\end{equation}
where $h_{v}^{(t)}$ is the node state at step $t$, $U_t\left(\cdot\right)$ is the update function and $m_{v}^{(t)}$ is the aggregated message at step $t$.

The message $m_{v}^{(t)}$ is defined as an aggregation of the neighbouring node states at step $t$:
\begin{equation}
    m_{v}^{(t)} = \sum_{w \in N(v)}^{} M_{t}\left(h_{v}^{(t)}, h_{w}^{(t)}, e_{vw}\right),
\end{equation}
where $M_t\left(\cdot\right)$ is the message function, which operates on each pair of nodes that share an edge $e_{vw}$.

Under the assumptions that message aggregation is computed as an adjacency-dependent sum, that the update function is based on linear combination of features and that both the message and update functions do not depend on $t$, we can reformulate the message passing problem in terms more akin to neural network layers: input features at each layer $t$ consist of the matrix of all node states $H^{(t)} = [ h_{1}^{(t)} ; h_{2}^{(t)} ; \dots ; h_{N}^{(t)}]$ ($N$ being the number of nodes in the graph), and the output of each layer $H^{(t+1)}$ is a function $f$ of the previous layer $H^{(t)}$ and the input graph adjacency matrix $A$:
\begin{equation}
    H^{(t+1)} = f(H^{(t)}, A).
\end{equation}

Following~\cite{gcn}, we add self loops to $A$ in order to take a node's own features into account when evaluating the corresponding message, and normalize the resulting matrix w.r.t. the diagonal node degree matrix:
\begin{equation}
    f(H^{(t)}, A) = a\left(\hat{D}^{-\frac{1}{2}}\hat{A}\hat{D}^{-\frac{1}{2}}H^{(t)}W^{(t)}\right),
\end{equation}
where $a\left(\cdot\right)$ is an activation function (ReLU in our case), $\hat{A} = A + I$ is the adjacency matrix with self loops, $\hat{D}$ is the diagonal node degree matrix, where each element is the row-wise sum of elements in $\hat{A}$, and $W^{t}$ is the weight matrix for layer $t$. Intuitively, each node in a GCN layer normalizes and aggregates features of adjacent nodes (including itself), before computing a new set of features through linear combination with the weight matrix followed by the activation function. 

In our approach, initial node features are provided as a learned embedding of object labels. We similarly model relationships associated to graph edges, by learning a corresponding embedding from relationship labels. Following \cite{you2020design}, we inject relationship information into the state update function by summing the relationship embeddings to node embeddings before aggregation.

\subsection{Autoregressive Image Generation}

Encoder-decoder transformers have shown outstanding performance in translation tasks~\cite{transformer},  thanks to their capability to capture global context through self-attention mechanisms and to find correlations at the token level between input and output sequences through cross-attention mechanisms.

We pose the image generation problem as a translation task between structured domain semantics, encoded by scene graphs, and the visual domain.

As shown in Fig.~\ref{fig:architecture}, the encoder module of the transformer receives as input the set of graph node representations, coming from the upstream GCN, applies standard positional encoding as in~\cite{transformer}, and feeds the resulting features to a series of encoder layers performing multi-head attention. Each layer has the same composition of the original transformer~\cite{transformer} and performs the following operations (in order):
\begin{equation}
    y_\text{MH} = \text{LayerNorm}(x + \text{MultiHead}(x)),
\end{equation}
\begin{equation}
    y = \text{LayerNorm}(y_\text{MH} + \text{FFN}(y_\text{MH})),
\end{equation}
where $x$ are features coming either from the previous attention layer or from the input embedding, \emph{LayerNorm} is layer normalization~\cite{layernorm}, and \emph{FFN} implements a feed-forward neural network.

\emph{MultiHead} represents multi-head attention, i.e., a linear projection over the encoder representation space of the concatenation of the outputs of multiple attention heads: 
\begin{equation}
    \text{MultiHead}(x) = \text{Concat}(\text{head}_1(x), \dots, \text{head}_h(x)) \cdot W^O,
\end{equation}

where $W^O$ is a learnable parameter matrix, and \emph{head}${}_i$ is:

\begin{equation}
    \text{head}_i(x) = \text{Attention}(xW^Q_i, xW^K_i, xW^V_i),
\end{equation}

$W^Q_i$, $W^K_i$, $W^V_i$ are projection matrices. The same $x$ is used for projection over query, key and values, thus implementing a \emph{self-attention} mechanism.

Finally, \emph{Attention} is computed as:

\begin{equation}
    \text{Attention}(Q, K, V) = \text{softmax}(\frac{QK^T}{\sqrt{d_k}})V.
\end{equation}

The transformer decoder has the objective of autoregressively generating the output image, i.e., by gradually adding scene elements as shown in Fig.~\ref{fig:decoder}.

At each generation step, the decoder receives the partial image generated up to that point, and produces a new updated image. At the beginning of the generation, the input can be either a blank image or a partial rendering of the target scene graph. Input images are compressed into a suitable latent space by means of a feature extractor; similarly, output features computed by the transformer decoder are projected back onto the pixel space by a generator module. The feature extractor and generator are jointly trained through an autoencoder, as described in the next section.

The internal architecture is similar to the encoder's, with two main differences. First, an additional \emph{cross-attention} operation is carried out in each decoder layer, where query vectors are computed from the decoder's outputs at the previous step, while key and value vectors are computed from the encoder's outputs. 
Second, masked self-attention is employed to prevent the decoder to exploit the knowledge of future elements in the sequence.

\begin{figure}
	\centering
	\includegraphics[width=0.5\textwidth]{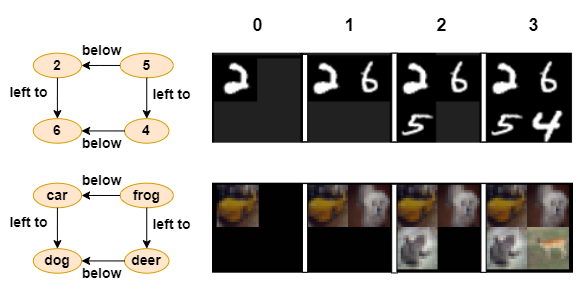}
	\caption{Sample ground truth sequences (scene graph and partial renderings) for our custom datasets \emph{Grid-MNIST} (above) and \emph{Grid-CIFAR} (below).
	}
	\label{fig:ground_truth}
\end{figure}

\subsection{Visual Representation Learning}

At each decoding step, the current sequence of partially generated input is provided to the decoder, which, in turn, is expected to extend that sequence with a new output. However, working directly with images in the pixel space poses several issues, not only in terms of hardware resources, but also because of the additional burden laid on the transformer to learn a suitable image representation.

To address these issues, we decouple the image representation problem from the joint graph-visual modeling, by separately training a convolutional autoencoder (AE) on the image dataset. The intermediate representation learned at the autoencoder bottleneck is therefore used as image features by the transformer, and the autoencoder's encoding and decoding paths as feature extractor and generator modules, respectively.

%% file: Sections/4_Results.tex
\begin{table*}
\centering
\caption{Results on \emph{Grid-MNIST} and \emph{Grid-CIFAR} of our full model, and some baselines}\label{table:sota}
\begin{tabular}{lllllllll}
\toprule
  & \multicolumn{4}{c}{Grid-MNIST} & \multicolumn{4}{c}{Grid-CIFAR10} \\
  \cmidrule(lr){2-5} \cmidrule(lr){6-9}
  & MSE ($\downarrow$) & SSIM ($\uparrow$) & MS-SSIM ($\uparrow$) & FID ($\downarrow$) & MSE ($\downarrow$) & SSIM ($\uparrow$) & MS-SSIM ($\uparrow$) & FID ($\downarrow$) \\
\midrule
Baseline 1 & 0.125 & 0.792 & 0.85 & 8.79 & 0.249 & 0.742 & 0.82 & 10.78 \\
Baseline 2 & \bfseries 4e-5 & \bfseries 0.999 & \bfseries 0.999 & 4.23 & \bfseries 7e-5 & \bfseries 0.999 & \bfseries 0.999 & \bfseries 4.76 \\
\midrule
\emph{sg2im}~\cite{johnson2018image}  &  6e-5 & \bfseries 0.999 & 0.993 & 5.86 & 1e-4 & 0.998 & 0.992 & 6.68\\
\emph{PasteGAN}~\cite{li2019pastegan}  &  5e-5 & \bfseries 0.999 & 0.995 & 4.89 & 8e-5 & 0.998 & 0.994 & 5.12\\
\midrule
Full Model & \bfseries 4e-5 & \bfseries 0.999 & 0.998 & \bfseries 4.21 & 0.001 &  0.990 & \bfseries 0.999 & 4.77 \\
\bottomrule
\end{tabular}
\end{table*}

\subsection{Dataset}

While the proposed approach is general, as a first proof-of-concept and without losing generality, we carry out our analysis on a simpler task than generating high-resolution natural images. 
We therefore built two custom datasets, \emph{Grid-MNIST} and \emph{Grid-CIFAR10}, respectively based on MNIST and CIFAR10, by randomly positioning source images in a N$\times$M grid and, accordingly, generating the related scene graphs. Each entry in the resulting datasets consists of: the final output image containing $N \times M$ images positioned in a grid, the related scene graph which describes the relative positioning of images in the grid, and $N \times M - 1$ intermediate grids (a set of partial images, obtained by adding a scene graph element at a time), which are used as the expected outputs at each generation step by the transformer. An example of a dataset item, for each dataset is shown in Fig.~\ref{fig:ground_truth}.

For comparison to baselines and state-of-the-art models, as well as for ablation studies, we use grids of 2$\times$2 images, each one resized to 16$\times$16 pixels, for a total grid size of 32$\times$32. 
As start-of-sequence token, we use a black image with the same size of the whole grid. Overall, each synthetic dataset consists of 10,000 grid images, with a 70/10/20 split into training, validation and test set.

\subsection{Training procedure and implementation details}
\label{subsec:training}

At training time we optimize an aggregated mean square error (MSE) objective between synthesized images and the corresponding ground truths, at each generation step. We used Adam as optimizer, with a learning rate of 10$^{-4}$ and a batch size of 64. Training is carried out for 200 epochs.

Our scene graph modeling network consists of three GCN layers, where the size of node and relationship embeddings is 64.
Transformer encoder and decoder apply four attention layers, each with four attention heads. Feature size for self- and cross-attention in the transformer is 256, while the output of feedforward layers is 2048.

It should be noted that the transformer module has a different behavior depending on whether we are in training or inference phase. In the first case, ``teacher forcing'' is carried out, i.e., all ground truth images are passed as input in parallel to the decoder (with proper attention masking), regardless of its intermediate outputs; at inference time, the decoder receives a start-of-sequence token only, with the following inputs autoregressively concatenated from previous outputs.

The autoencoder for image representation learning is based on ResNet-18~\cite{he2015deep} in both of its encoding and decoding paths, with the latter replacing convolutions and pooling with transposed convolutions and bilinear upsampling. Feature size at bottleneck is 256. The autoencoder is the only pre-trained (on the target custom dataset) component of the architecture, while the GCN and transformer are trained end-to-end.

\subsection{Results}

We evaluate the correctness and reconstruction quality of our generation approach from scene graphs, by computing MSE and SSIM~\cite{wang2004image} scores to measure the distance between pixel values and the similarity between predicted and ground truth image structure, at the final generation step. We also compute MS-SSIM~\cite{wang2003multiscale} to give an evaluation on the image structure, and FID~\cite{fid} to assess coherence between generated and ground-truth image features.
We quantitatively compare our results to \emph{sg2im}~\cite{johnson2018image}, \emph{PasteGAN}~\cite{li2019pastegan} (as they employ scene graphs for synthesis) and two additional baselines: \emph{Baseline 1} combines the GCN and the generator network from our model, using the sum-based graph readout function~\cite{micheli2009neural} to condition generation; \emph{Baseline 2} combines our GCN and transformer encoder to learn node embeddings (again aggregated through sum readout) to condition the generator network.
Results are shown in Table~\ref{table:sota}, and demonstrate that graph modeling alone is not sufficient to condition the generation process, which requires either a transformer encoder or an ad-hoc convolutional framework as in~\cite{johnson2018image}. 

\begin{figure}
	\centering
	\includegraphics[width=0.5\textwidth]{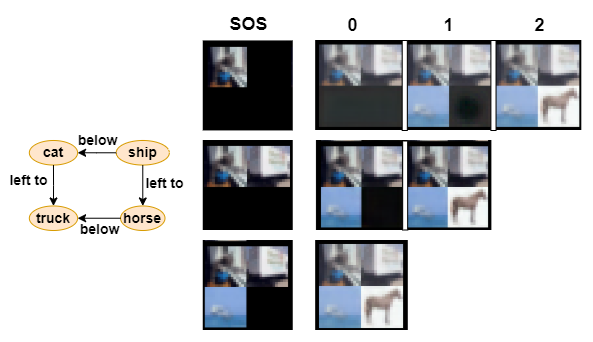}
	\caption{\textbf{Examples of image generation from partial images.} Left: input scene graph. Center: partial renderings used as start-of-sequence to the transformer decoder; from top to bottom, generation with one, two or three starting elements. Right: transformer outputs at each decoding step.}
	\label{fig:partial_imgs}
\end{figure}

Table~\ref{table:sota} shows that, given enough representational power, all methods under analysis correctly synthesize images corresponding to input scene graphs with no significant quantitative difference; however, our model is the only one that supports generation from partial renderings. Fig.~\ref{fig:partial_imgs} shows examples of images generated from the same scene graph by providing different initial conditions to the transformer encoder: the model successfully fills missing elements with the corresponding objects from the scene graph specification.

\begin{figure}
	\centering
	\includegraphics[width=0.45\textwidth]{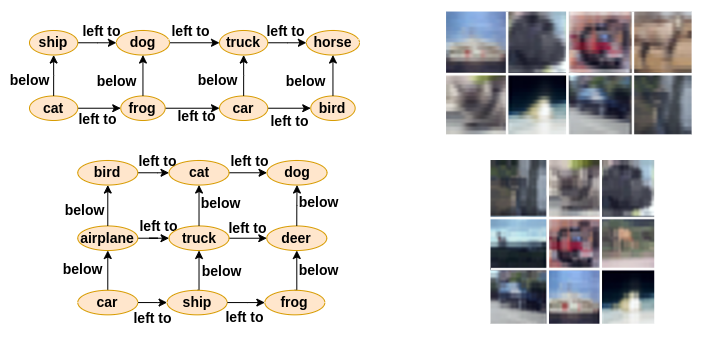}
	\caption{\textbf{More examples of generation.} Left: input scene graphs. Right: final output of the model for the correspondent scene graph.}
	\label{fig:complex_graphs}
\end{figure}

In order to substantiate our design choices, we perform an ablation study that, starting from the architecture used as \emph{Baseline 1}, quantitatively evaluates the impact of adding, in order, the transformer encoder, the transformer decoder and the autoencoder for visual feature learning. For the first three configurations (i.e., without the autoencoder), we directly use flattened pixel values as the model's input and output. Results for both datasets are shown in Table~\ref{table:abl}, highlighting that the use of a transformer architecture significantly improves the accuracy of the generation.
It is interesting to note that, while the use of the autoencoder marginally improves performance, it has a significant impact on training convergence. When using flattened pixel values as input and output to the decoder (hidden size respectively of 1,024 for Grid-MNIST and 3,072 for Grid-CIFAR10), the transformer requires up to 1,000 training epochs to converge, compared to the 200 epochs sufficient to train the variant with the autoencoder.

\begin{table}
\centering
\caption{Ablation study on the Grid-CIFAR10 dataset, with standard deviations computed on five different runs.}\label{table:abl}
\begin{tabular}{lcc}
\toprule
  & MSE ($\downarrow$) & SSIM ($\uparrow$)\\
\midrule
GCN & 1.729  $\pm$ 0.068 & 0.083 $\pm$ 0.005 \\
\hspace{0.1cm} + Transf. encoder & 0.036 $\pm$ 0.003 & 0.938 $\pm$ 0.004 \\
\hspace{0.2cm} + Transf. decoder &  0.006 $\pm$ 0.005 & 0.983 $\pm$ 0.003 \\
\hspace{0.3cm} + Autoencoder &  {\bfseries 0.001 $\pm$ 0.002} & {\bfseries 0.991 $\pm$ 0.003} \\
\hline
\end{tabular}
\end{table}

We then carry out a set of experiments to identify the best configuration for the transformer model. Starting from our base empirical configuration that uses four attention layers and four attention heads, we measure the effect of changing the number of attention layers. Table~\ref{tab:ae_arch} shows that reducing the number of attention layers slightly worsens performance, while increasing it leads to severe overfitting. After setting the number of layers to 4, we change the number of attention heads in each attention layer. The same Table~\ref{tab:ae_arch} also shows that reducing the number of attention heads has a negative impact on performance, while increasing it does not seem to provide benefits, hence we opt for the simple configuration.

\begin{figure}
	\centering
	\includegraphics[width=0.42\textwidth]{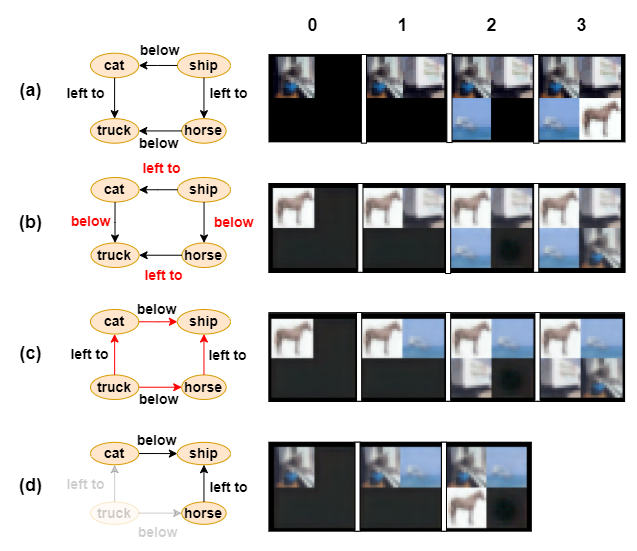}
	\caption{Examples of the effect of graph modifications on the generated image: (a) original graph; (b) changed relationship embeddings; (c) changed relationship directions; (d) removed graph node.}
	\label{fig:alterations}
\end{figure}

Finally, we perform a qualitative analysis on how the model reflects variations in the input scene graph onto generated outputs.
Fig.~\ref{fig:complex_graphs} reports the behavior of the model with more complex graphs, which involve more objects and relationships than the simpler 2$\times$2 examples.
Fig.~\ref{fig:alterations} shows the effect of several modifications applied to an example scene graph: changing relationships between nodes, reversing relationship directions, removing a node from the graph. In all of these cases, our model produces images that reflect changes in the underlying scene graph, demonstrating its capability to capture a variety of object configurations. 

\begin{table}
\centering
\caption{Effect of the number N of attention layers and heads in the transformer model}\label{tab:ae_arch}
\begin{tabular}{lllll}
\toprule
 & \multicolumn{2}{c}{\textbf{Attention Layers}} & \multicolumn{2}{c}{\textbf{Attention Heads}} \\
  \cmidrule(lr){2-3} \cmidrule(lr){4-5} 
 $N$ & MSE ($\downarrow$) & SSIM & MSE ($\downarrow$) & SSIM ($\uparrow$)\\
\midrule

2 & 0.035  & 0.899 & 0.149 & 0.861 \\
4 & \bfseries 0.009 & \bfseries 0.985 & \bfseries 0.009 &  0.989 \\
8 & 0.547 & 0.580 & 0.010 & \bfseries 0.990 \\

\bottomrule
\end{tabular}
\end{table}

%% file: Sections/5_Conclusions.tex
We propose an autoregressive approach based on encoder-decoder transformers for image generation conditioned by scene graphs. We show that, aside from the ability to compose an image by gradually introducing elements from the scene graph into the generated image in a semantically consistent way, our model is capable of satisfying scene constraints by filling a partially generated image provided as input, introducing a novel, user-friendly way to condition the generation process. Our results show that our method reproduces objects in the scene in a visually-accurate way and is able to handle a variety of configurations between objects in the graph.
To the best of our knowledge, our work presents a yet unexplored approach to employ transformers for joint semantic-driven and user-conditioned image generation. The obtained results are generated in simple settings that we aim to move to the more challenging Visual Genome \cite{visualgenome} dataset. 

%% file: Sections/6_Acknowledgements.tex
This work has been partially funded by the REHASTART project - Regione Sicilia (PO FESR 2014/2020 - Azione 1.1.5, N. 08ME6201000222,CUP  G79J18000610007) and by Piano della Ricerca di Ateneo 2020-2022–PIACERI Starting Grant Università di Catania, Project: BrAIn: Adaptive Brain-Derived Artificial Intelligence Methods for Event Detection.